\documentclass[letterpaper]{article} 
\usepackage{aaai24}  
\usepackage{times}  
\usepackage{helvet}  
\usepackage{courier}  
\usepackage[hyphens]{url}  
\usepackage{graphicx} 
\usepackage{natbib}  
\usepackage{caption} 
\usepackage{algorithm}
\usepackage{algorithmic}

\usepackage{newfloat}
\usepackage{listings}
\DeclareCaptionStyle{ruled}{labelfont=normalfont,labelsep=colon,strut=off} 
\floatstyle{ruled}
\newfloat{listing}{tb}{lst}{}
\floatname{listing}{Listing}
\usepackage{amsthm}
\usepackage{stmaryrd}
\usepackage{amsmath}
\usepackage{amssymb}
\usepackage{siunitx}
\usepackage{bm}

\usepackage{graphicx}
\usepackage{caption}
\usepackage{subcaption}
\usepackage[table]{xcolor}
\usepackage{color}
\usepackage{svg}

\usepackage{multirow}
\usepackage{makecell} 
\usepackage{booktabs}

\usepackage{lipsum} 
\usepackage{pdfpages}
\usepackage{kotex}
\usepackage{dsfont}

\DeclareMathOperator*{\argmax}{argmax}

\title{LINGO-Space: Language-Conditioned Incremental Grounding for Space}
\author{
    Dohyun Kim,
    Nayoung Oh,
    Deokmin Hwang,
    Daehyung Park\thanks{D. Park is the corresponding author.}
}
\affiliations{
    Korea Advanced Institute of Science and Technology, Republic of Korea\\

    \{dohyun141, lightsalt, gsh04089, daehyung\}@kaist.ac.kr
}

\begin{document}

\maketitle

\begin{abstract}
We aim to solve the problem of spatially localizing composite instructions referring to space: space grounding. Compared to current instance grounding, space grounding is challenging due to the ill-posedness of identifying locations referred to by discrete expressions and the compositional ambiguity of referring expressions. Therefore, we propose a novel probabilistic space-grounding methodology (LINGO-Space) that accurately identifies a probabilistic distribution of space being referred to and incrementally updates it, given subsequent referring expressions leveraging configurable polar distributions. Our evaluations show that the estimation using polar distributions enables a robot to ground locations successfully through $20$ table-top manipulation benchmark tests. We also show that updating the distribution helps the grounding method accurately narrow the referring space. We finally demonstrate the robustness of the space grounding with simulated manipulation and real quadruped robot navigation tasks. Code and videos are available at https://lingo-space.github.io.
\end{abstract}
\section{Introduction}
Robotic natural-language grounding methods have primarily focused on identifying objects, actions, or events~\cite{brohan2023can, mees23hulc2}. However, to robustly carry out tasks following human instructions in physical space, robots need to identify the space of operations and interpret spatial relations within the instructions. For example, given a directional expression (e.g., ``place a cup on the table and close to the plate"), a robot should determine the most suitable location for placement based on the description and environmental observations.

\begin{figure}[t]
  \centering
  \includegraphics[width=0.9\columnwidth]{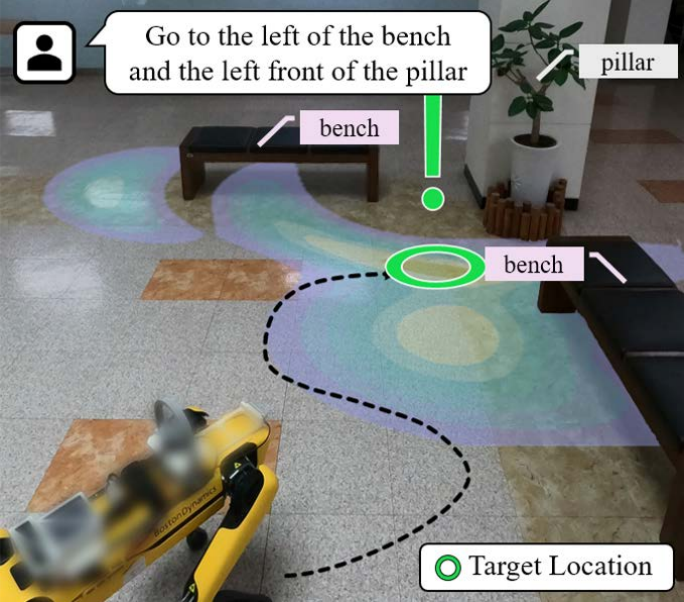}
  \caption{An illustration of incremental \textit{space grounding}  in the navigation task. Our method, LINGO-Space, identifies the distribution of the target location indicated by a natural language instruction with referring expressions.}
  \label{fig_main}
\end{figure}

We aim to solve the problem of localizing spatial references within instructions. Referred to as ``space grounding," this problem involves identifying potential locations for reaching or placing objects. Unlike conventional \textit{instance grounding} problems, such as visual object grounding~\cite{shridhar2020ingress} and scene-graph grounding~\cite{kim2023sggnet}, space grounding presents complexities due to inherent ambiguity in identifying referred locations (as illustrated in Fig.~\ref{fig_main}). Given the ambiguity and the compositional nature of referring expressions, a grounding solution should be capable of reasoning about potential space candidates with uncertainty and adapting to new references. 

Conventional space grounding approaches often map spatially relational expressions (e.g., ``to the right side of a box") to specific directional and distance-based coordinates, learning patterns from training dataset~\cite{jain2023ground, namasivayam2023learning}. However, \textit{positional ambiguity} (e.g., missing distance information) in spatial expressions and \textit{referential ambiguity} (e.g., a plurality of similar objects) in the scene often lower effective space grounding~\cite{zhao2023differentiable}. Furthermore, \textit{representational ambiguity} restricts the scalability of space grounding when dealing with complex expressions and scenes.

On the other hand, understanding composite expressions is crucial for accurate grounding in space. Most instructions entail sequences of spatiotemporal descriptions (e.g., ``Enter the room, then place the cup on the table"). However, most approaches often encode multiple expressions simultaneously without explicit separation~\cite{roy2019leveraging}. Recently, \textsc{ParaGon}~\cite{zhao2023differentiable} decomposes a composite instruction into object-centric relation expressions and jointly encodes them using graph neural networks. Given the variability in results due to the expression order and incoming expressions, we need an incremental grounding method with composite expressions.

Therefore, we propose LINGO-Space, a language-conditioned incremental grounding method for space.\footnote{LINGO-Space is an abbreviation of \textbf{L}anguage-conditioned \textbf{IN}cremental \textbf{G}r\textbf{O}unding method for \textbf{Space}} This method identifies a probabilistic distribution of the referenced space by leveraging configurable polar distributions. Our method incrementally updates the distribution given subsequent referring expressions, resolving \textit{compositional ambiguity} via a large language model (LLM)-guided semantic parser. We also mitigate \textit{referential ambiguity} by leveraging scene-graph-based representations in grounding.

Our evaluation shows that estimating polar distributions effectively grounds space as described by referring expressions, while conventional methods have difficulty capturing uncertainty. We also show the capability to refine the distribution accurately and narrow down the referenced space as humans encounter space navigation in complex domains.

Our contributions are as follows:
\begin{itemize}
    \item We propose a novel space representation using a mixture of configurable polar distributions, offering a probability distribution of referred locations.
    \item We introduce an incremental grounding network integrated with an LLM-based semantic parser, enabling robust and precise grounding of incoming expressions.
    \item We conduct $20$ benchmark evaluations, comparing with state-of-the-art baselines, and demonstrate the real-world applicability of our method through space-reaching experiments involving a quadruped robot, Spot.
\end{itemize}
\section{Related Work}
\noindent\textbf{Language grounding}: The problem of language instruction delivery has received increasing attention in robotics. Early works have focused on understanding entities or supplementary visual concepts \cite{matuszek2012joint}. 
Recent works have incorporated spatial relations to enhance the identification of instances referred to in expressions \cite{howard2014natural, paul2018efficient, hatori2018interactively, shridhar2020ingress}. 

\noindent\textbf{Space grounding}: There are efforts mapping spatial relations to the region of actions using various representations: potential fields~\cite{stopp1994utilizing}, discrete regions with fuzzy memberships~\cite{tan2014grounding}, and points from a multi-class logistic classifier~\cite{guadarrama2013grounding}. Early approaches have employed predetermined distances or directions, or randomly sampled locations to represent spatial relations. Neural representations have emerged predicting pixel positions~\cite{venkatesh2021spatial} or pixel-wise probabilities~\cite{mees2020learning} for placement tasks.  
To overcome limitations associated with pixel-based distributions, researchers have used parametric probability distributions, such as a polar distribution~\cite{kartmann2020representing}, a mixture of Gaussian distributions~\cite{zhao2023differentiable}, and a Boltzmann distribution~\cite{gkanatsios2023energy}. Our proposed method adopts the polar distribution as a basis for modeling spatial concepts, avoiding the need to predefine the number of components as required by Gaussian mixture models~\cite{kartmann2020representing, paxton2022predicting}. Further, our method considers the order of expressions and the semantic and geometric relations among objects, allowing for handling semantically identical objects. 

\noindent\textbf{Composite instructions}: Composite linguistic instructions often introduce \textit{structural ambiguity}. Researchers often use \textit{parsing} as a solution, breaking down composite expressions using hand-crafted or grammatical rules~\cite{tellex2011understanding, howard2022intelligence}. Recently, \citet{zhao2023differentiable} have introduced the grounding method, \textsc{ParaGon}, in which its neural parsing module extracts object-centric relations. Similarly, \citet{gkanatsios2023energy} decompose expressions into spatial predicates using a neural semantic parser, i.e., a sequence-to-tree model~\cite{dong2016language}. While these works often deal with one or two referring expressions, our method incrementally manages an arbitrary number of referring expressions by introducing an LLM-based parser.

\noindent\textbf{Large language models}: LLMs have brought increasing attention offering benefits in the areas of understanding high-level commands~\cite{brohan2023can}, extracting common manipulation-related knowledge~\cite{ren2023leveraging}, planning with natural language commands~\cite{huang2022inner, song2023llm, driess2023palme, mees23hulc2}, and programming~\cite{singh2023progprompt, liang2023code}. While these approaches generally focus on the LLM's capability to leverage semantic knowledge for understanding natural language instructions, we focus on another capability: decomposing linguistically complex commands into sub-commands or problems. \citet{shah2022lmnav} employ an LLM to generate a list of landmarks within composite commands. Similarly, \citet{liu2023lang2ltl} use an LLM to identify referring expressions and translate natural language commands into linear temporal logics. Our method also uses an LLM to parse composite referring instructions and transform them into a structured format, enhancing the grounding process.
\section{Problem Formulation}

\begin{figure*}[t]
  \centering
  \includegraphics[width=0.95\textwidth]{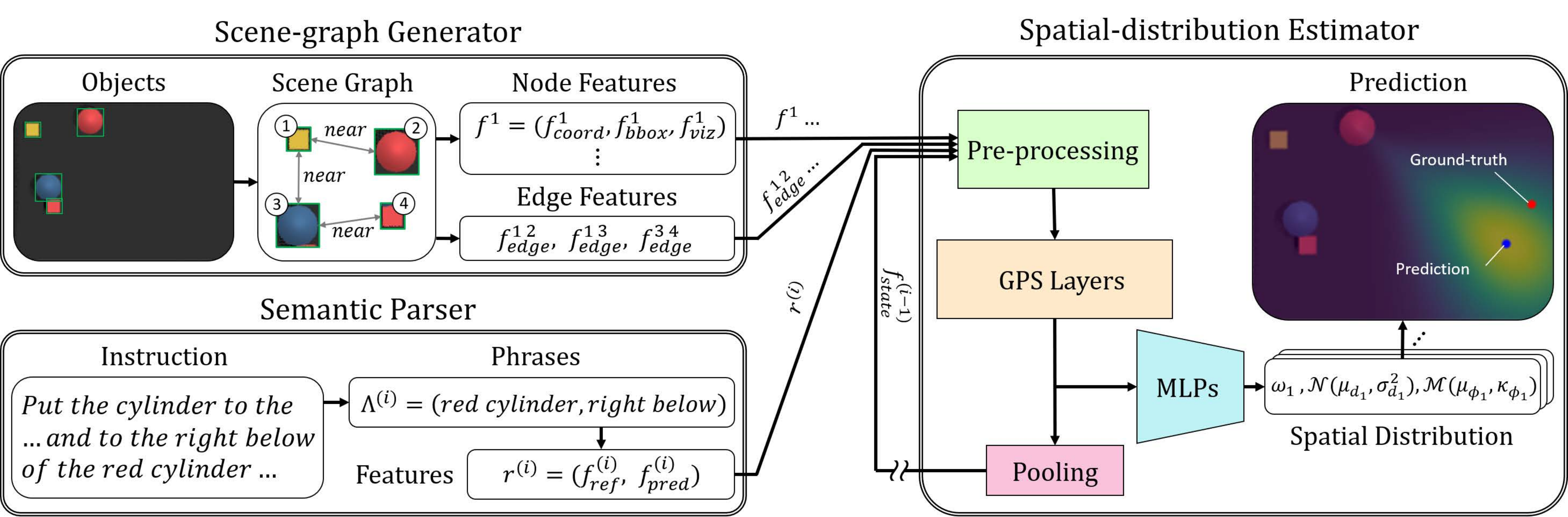}
  \caption{The overall architecture of LINGO-Space on a tabletop manipulation task. Given a composite instruction, a graph generator provides a scene graph. A semantic parser decomposes the instruction into a structured form of relation-embedding tuples $r^{(i)}$, where $i\in\{1, ..., M\}$. Finally, a spatial-distribution estimator incrementally updates a probabilistic distribution of locations satisfying spatial constraints encoded in the embedding tuples.}
  \label{fig:pipeline}
\end{figure*}

Consider the problem of determining a desired location $\mathbf{x}^* \in \mathbb{R}^2$ based on a natural language instruction $\Lambda$, while taking into account a set of objects $\mathcal{O}=\{o_1, ... o_N \}$ in the environment, where $N$ denotes the number of objects. To enhance the accuracy of indicating the location $\mathbf{x}^*$, the instruction $\Lambda$ might include object references that leverage their geometric relationships with the intended location. To robustly represent potential locations, referred to as ``space," we model the target location as a probability distribution parameterized by $\bm{\theta}$. Therefore, we formulate an optimization problem in which we marginalize out $\bm{\theta}$ and a scene graph $\Upsilon_{\text{sg}}$ that encodes $\mathcal{O}$ with their relationships (i.e., edge features):
\begin{align}
\mathbf{x}^* &= \argmax_{\mathbf{x}} P(\mathbf{x} | \Lambda, \mathcal{O}), \\
&= \argmax_{\mathbf{x}} \iint_{\bm{\theta}, \Upsilon_{\text{sg}}}  P(\mathbf{x}|\bm{\theta}) P(\bm{\theta} | \Lambda, \Upsilon_{\text{sg}}) P(\Upsilon_{\text{sg}} | \mathcal{O} ),
\label{eq_problem}
\end{align}
where we assume conditional independence between the current distribution model $\bm{\theta}$, given the scene graph $\Upsilon_{\text{sg}}$, and the object set $\mathcal{O}$. This work assumes a scene-graph generator produces an optimal graph $\Upsilon_{\text{sg}}^*$.

Another problem is the use of composite instructions with multiple relations (i.e., referring expressions) in sequence. We assume a plurality of similar reference objects to be present in the environment, potentially leading to instances where semantically identical labels appear on the graph $\Upsilon_{\text{sg}}^*$. To mitigate compositional ambiguity in composite instructions, we decompose $\Lambda$ into multiple phrases, $\Lambda=[\Lambda^{(1)}, ... , \Lambda^{(M)}]$, where $M$ is the number of constituent phrases. Each phrase $\Lambda^{(i)}$ includes a single spatial relation about a referenced object (e.g., ``left of the block''). We then reformulate the objective function in Eq.~(\ref{eq_problem}) using an iterative update form with $\Upsilon_{\text{sg}}^*$:
\begin{align}
\underbrace{P(\mathbf{x}|\bm{\theta}_M)}_{\substack{\text{Location}\\ \text{selector}}} \prod_{i=1}^M [\underbrace{P(\bm{\theta}_i | \bm{\theta}_{i-1}, \Lambda^{(i)}, \Upsilon_{\text{sg}}^*)}_{\substack{\text{Spatial-distribution}\\ \text{estimator}}} \underbrace{P(\Lambda^{(i)} |\Lambda, \Upsilon_{\text{sg}}^*)}_{\text{Semantic parser}}],
\label{eq_iter_problem}
\end{align}
where $P(\bm{\theta}_1 | \bm{\theta_0}, \Lambda^{(1)}, \Upsilon_{\text{sg}})=P(\bm{\theta}_1 | \Lambda^{(1)}, \Upsilon_{\text{sg}})$.
We describe each process and the graph generator below.
\section{Methodology: LINGO-Space}
We present LINGO-Space, an incremental probabilistic grounding approach that predicts the spatial distribution of the target space referenced in a composite instruction. The architecture of our method, as depicted in Fig.~\ref{fig:pipeline}, consists of 1) a scene-graph generator, 2) a semantic parser, and 3) a spatial-distribution estimator. Below, we describe each module and the incremental process of estimating the spatial distribution to ground the desired location effectively.

\subsection{A Scene-Graph Generator}
A scene graph $\Upsilon_\text{sg} = (\mathcal{V}, \mathcal{E})$ is a graphical representation of a scene consisting of detected objects as nodes $\mathcal{V}$  and their pairwise relationships as directed edges $\mathcal{E}$.
Each node $u \in\mathcal{V}$ entails node features $\mathbf{f}^u$ including its Cartesian coordinate $\mathbf{f}^u_{\text{coord}}\in\mathbb{R}^2$, bounding box $\mathbf{f}^u_{\text{box}}\in\mathbb{R}^{4}$, and visual feature $\mathbf{f}^u_{\text{viz}}\in \mathbb{R}^{D_{\text{viz}}}$; $\mathbf{f}^u=(\mathbf{f}^u_{\text{coord}}, \mathbf{f}^u_{\text{box}}, \mathbf{f}^u_{\text{viz}})$, where $D_{\text{viz}}$ is a fixed size. In detail, a bounding box detector (e.g., Grounding DINO~\cite{liu2023grounding} for manipulation) finds $\mathbf{f}^u_{\text{coord}}$ and $\mathbf{f}^u_{\text{box}}$. Then, we encode its cropped object image as $\mathbf{f}^u_{\text{viz}}$ using the CLIP image encoder~\cite{radford2021learning}.
Each edge $e_{uv}\in\mathcal{E}$ represents a spatial relationship, as a textual predicate, from $u$ to $v\in\mathcal{V}$. We determine predicates (i.e., ``near", ``in") using the box coordinate and size, ($\mathbf{f}_{\text{coord}}, \mathbf{f}_{\text{box}}$). 
We encode each predicate as an edge feature $\mathbf{f}^{uv}_{\text{edge}}\in \mathbb{R}^{D_{\text{txt}}}$ using the CLIP text encoder~\cite{radford2021learning}, where $D_{\text{txt}}$ is a fixed size.

We design a scene-graph generator that returns a graph $\Upsilon_{\text{sg}}$ in a dictionary form; $\mathcal{V}$ is a dictionary with node identification numbers (ID) as keys and node features as values, e.g., \{$23$: [$\mathbf{f}^{23}_{\text{coord}}, \mathbf{f}^{23}_{\text{box}}, \mathbf{f}^{23}_{\text{viz}}$] \}, where $23$ is an ID number. Note that we use node IDs as an interchangeable concept with nodes. We also represent an edge set $\mathcal{E}$ as a list of relationship triplets, ($u_{\text{ID}},\mathbf{f}^{u_{\text{ID}} v_{\text{ID}}}_{\text{edge}},v_{\text{ID}}$), where $u_{\text{ID}}$ and $v_{\text{ID}}$ are the subject and object node IDs. This work assumes that each edge contains only one relation in a pre-defined set.

\subsection{A Semantic Parser}
We introduce an LLM-based semantic parser, using ChatGPT~\cite{openai2023chatgpt}, which 1) breaks down a composite instruction $\Lambda$ with $M$ referring expressions into its constituent phrases $[\Lambda^{(\text{main})}, \Lambda^{(1)}, ..., \Lambda^{(M)}]$ and 2) transforms these phrases into a structured format, leveraging the LLM's proficiency in in-context learning through prompts. We design the prompt to consist of a task description and parsing demonstrations.
The task description explains our parsing task as well as its reasoning steps: 1) identification of an \textit{action} from the instruction, 2) identification of a \textit{source} instance associated with the \textit{action}, and 3) identification of \textit{target} information from the referring expressions, characterized by a relational predicate and a referenced object. Then, the demonstrations provide three input-output examples to regularize the output format.

In detail, our parser represents the main phrase $\Lambda^{(\text{main})}$ as an \textit{action} with an associated \textit{source} object and the other referring phrases, $[\Lambda^{(1)}, ..., \Lambda^{(M)}]$, as relational predicates with referenced objects. To incrementally ground the phrases, our parser converts the referenced object-predicate pairs as a list of relation tuples, $[r^{(1)}, ..., r^{(M)}]$. For instance,
\noindent\fbox{
    \parbox{0.95\linewidth}{
        \textbf{Input}: put the cyan bowl above the chocolate and left of the silver spoon. \\
        \textbf{Output}: \{action: ``move'', source: ``cyan bowl'', target: [(``chocolate'', ``above''), (``silver spoon'', ``left'')] \}.
    }
}

We then post-process the output form to have better representations for grounding. For it, we replace the \textit{action} into a robot skill with the highest word similarity in a skill set.
We also replace the text-based $r^{(i)}$ into an embedding-based tuple $r^{(i)}=(\mathbf{f}_{\text{ref}}^{(i)}, \mathbf{f}_{\text{pred}}^{(i)})$, where $\mathbf{f}_{\text{ref}}^{(i)}\in\mathbb{R}^{D_{\text{txt}}}$ and $\mathbf{f}_{\text{pred}}^{(i)}\in\mathbb{R}^{D_{\text{txt}}}$ are encoded from the CLIP text encoder~\cite{radford2021learning}.

\begin{figure}[t]
  \centering
  \includegraphics[width=\columnwidth]{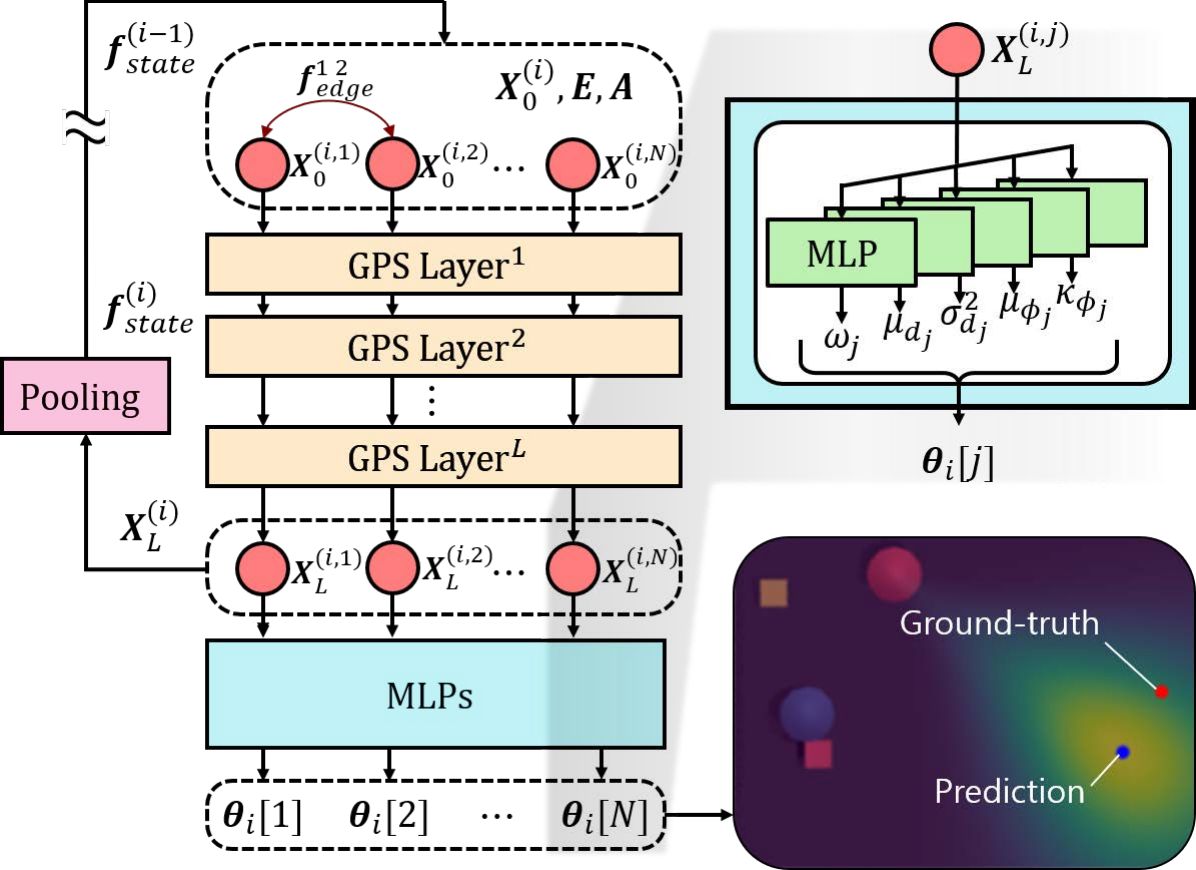}
  \caption{Architecture of the spatial-distribution estimator. Given the graph representation of the problem description, the network predicts instance-wise polar distributions, updating the internal model with the previous state.}
  \label{fig:estimation_model}
\end{figure}

\subsection{A Spatial-Distribution Estimator}
Our spatial distribution estimator predicts a probability distribution of destinations given a referring phrase and a scene graph. Given a sequence of phrases, the estimator incrementally updates the distribution using a graph-based incremental grounding network.

\subsubsection{Spatial distribution}
We represent the probability distribution as a mixture of instance-wise polar distributions, where a polar distribution is a joint probability density function of two random variables, distance $d\in\mathbb{R}_{\geq 0}$ and angle $\phi\in[-\pi, \pi]$, in the polar coordinate system. As \citet{kartmann2020representing}, we assume that $d$ and $\phi$ follow a Gaussian distribution $\mathcal{N}$ and a Von Mises (i.e., circular) distribution $\mathcal{M}$, respectively;
\begin{align}
(d, \phi) \sim (\mathcal{N}(\mu_d, \sigma^2_d), \mathcal{M}(\mu_\phi, \kappa_\phi) ),
\end{align}
where $\mu_d$, $\sigma_d^2$, $\mu_\phi$, and $\kappa_\phi$ indicate mean, variance, location, and concentration, respectively. Then, the mixture of instance-wise polar distribution given a tuple $r^{(i)}$ is,
\begin{align}
    P(\mathbf{x}| r^{(i)}, \Upsilon_{\text{sg}}) = \sum^{N}_{j=1} w_j \cdot P(d; \mu_{d_j}, \sigma_{d_j}^2) \cdot P(\phi; \mu_{\phi_j}, \kappa_{\phi_j}).
    \label{eq:weighted_sum}
\end{align}
where $w_j$ is a weight that represents the relevance between the $j$-th instance and the relation tuple $r^{(i)}$. We represent the entire distribution as the mixture model parameters $\bm{\theta}$: $\bm{\theta}=[(w_1, \mu_{d_1}, \sigma^2_{d_1}, \mu_{\phi_1}, \kappa_{\phi_1} ), ... , (w_N, \mu_{d_N}, \sigma^2_{d_N}, \mu_{\phi_N}, \kappa_{\phi_N} ) ]$.

\subsubsection{{Pre-processing}}
Given a node feature $\mathbf{f}^{j}$ and a relation tuple $r^{(i)}$, we pre-process them to have better representations by projecting into another space via \textit{positional encoding} and \textit{matrix projection} processes. The \textit{positional encoding} embeds the Cartesian coordinate $\mathbf{f}^{j}_{\text{coord}}$ using sinusoidal positional encoding $\gamma(\cdot)$: $\bar{\mathbf{f}}^{j}_{\text{coord}}=\gamma(\mathbf{f}^{j}_{\text{coord}})\in\mathbb{R}^{2(2K+1)}$ where $K$ is a predefined maximum frequency $K\in\mathbb{N}$~\cite{mildenhall2021nerf}. The \textit{matrix projection} process projects each feature (i.e., $\mathbf{f}^{j}_{\text{viz}}$, $\mathbf{f}^{(i)}_{\text{ref}}$, and $\mathbf{f}^{(i)}_{\text{pred}}$) into a new feature space with dimension $D_{H}$ by multiplying a learnable projection matrix:
\begin{align}
\bar{\mathbf{f}}^j_{\text{viz}}=\mathbf{M}_{\text{viz}}\mathbf{f}^j_{\text{viz}}, \quad \bar{\mathbf{f}}^{(i)}_{\text{ref}}=\mathbf{M}_{\text{ref}}\mathbf{f}^{(i)}_{\text{ref}}, \quad \bar{\mathbf{f}}^{(i)}_{\text{pred}}=\mathbf{M}_{\text{pred}}\mathbf{f}^{(i)}_{\text{pred}}, \nonumber
\end{align}
where $\mathbf{M}_{\text{viz}}\in\mathbb{R}^{D_H \times D_{\text{viz}}}$, $\mathbf{M}_{\text{ref}}\in\mathbb{R}^{D_H \times D_{\text{CLIP}}}$, and $\mathbf{M}_{\text{pred}}\in\mathbb{R}^{D_H \times D_{\text{CLIP}}}$. In addition, to use the last estimation model state $\mathbf{f}^{(i-1)}_{\text{state}}\in\mathbb{R}^{N\cdot D_{H'}}$, we also compress $\mathbf{f}^{(i-1)}_{\text{state}}$ into $\bar{\mathbf{f}}^{(i-1)}_{\text{state}}\in\mathbb{R}^{D_H}$ by applying a \textit{max-pooling} operation and a linear projection,
where $D_{H'}=4D_{H}+2(2K+1)$. Therefore, we obtain a new feature vector $\mathbf{X}^{(i,j)}_0 \in \mathbb{R}^{D_{H'}}$ that is a concatenation of the projected feature vectors,
\begin{align}
\mathbf{X}^{(i,j)}_0=
concat(\bar{\mathbf{f}}^{j}_{\text{coord}}, \bar{\mathbf{f}}^{j}_{\text{viz}}, \bar{\mathbf{f}}_{\text{ref}}^{(i)}, \bar{\mathbf{f}}_{\text{pred}}^{(i)}, \bar{\mathbf{f}}^{(i-1)}_{\text{state}} ).
\end{align}
For $i = 1$, we use a zero vector as the last model state.

\begin{figure*}[t]
  \centering
  \includegraphics[width=0.95\textwidth]{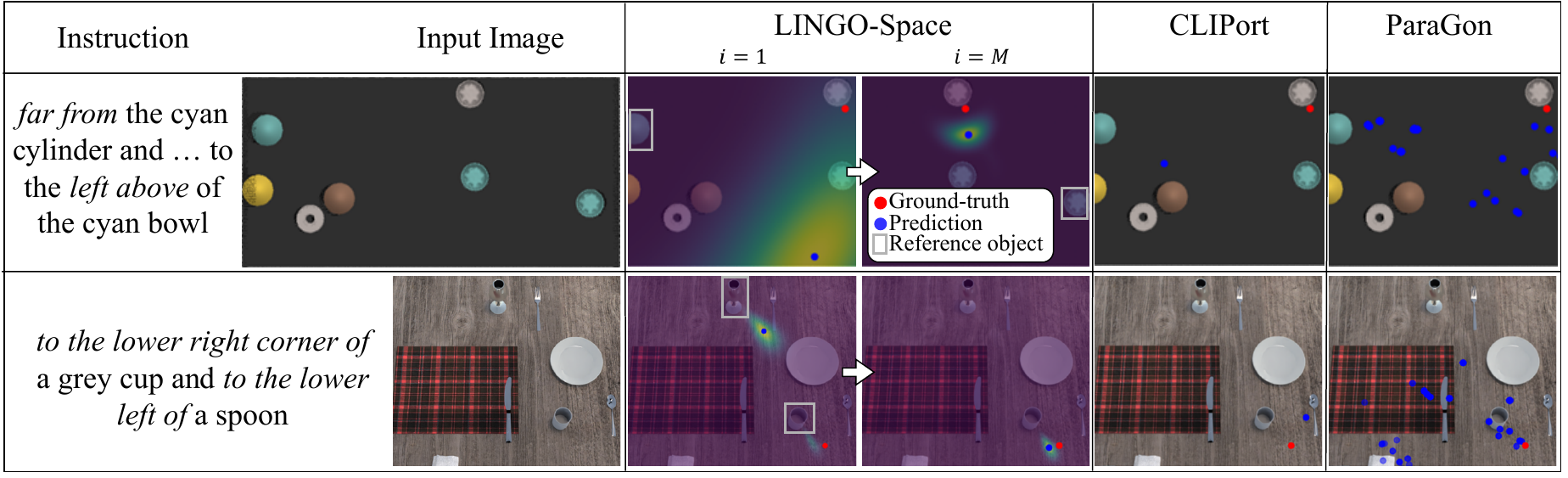}
  \caption{Qualitative evaluation with LINGO-Space, \textsc{CLIPort}, and \textsc{ParaGon}. Grey boxes represent the object each $i$-th phrase refers to, while red dots and blue dots represent the ground-truth and the prediction, respectively. We plot $100$ particles for the \textsc{ParaGon}'s prediction result. The results demonstrate that LINGO-Space is capable of accurately identifying the space referred to by a composite instruction by narrowing down the space.}
  \label{fig:qualitative_result}
\end{figure*}

\subsubsection{{Estimation network}}
We design the estimation network to predict instant-wise polar distributions given a relation tuple $r^{(i)}$, taking new node features $\mathbf{X}_0=(\mathbf{X}_0^{(i,1)}, ... , \mathbf{X}_0^{(i,N)})$ and edge features $\mathbf{E}_0=(... , \mathbf{f}^{uv}_{\text{edge}}, ...)$. Fig.~\ref{fig:estimation_model} shows the network architecture, which is a stack of GPS layers~\cite{rampavsek2022recipe}, where each GPS layer is a hybrid layer of a message-passing neural network (MPNN) and a global attention network. This work uses GINE~\cite{hu2020strategies} as the MPNN layer and Transformer~\cite{vaswani2017attention} as the global attention layer. The GPS layer allows the network to update node and edge features,
\begin{align}
\mathbf{X}_{l+1}, \mathbf{E}_{l+1} = \mathbf{GPS}^{l}(\mathbf{X}_{l}, \mathbf{E}_{l}, \mathbf{A}),
\end{align}
where $\mathbf{A}\in\mathbb{R}^{N\times N}$ is the adjacency matrix of the scene graph $\Upsilon_{\text{sg}}$, $l\in\{1, ..., L\}$, and $L$ is the number of layers.   

In the estimation network, the $l$-th GPS layer returns instant-wise hidden states $\mathbf{X}_{l}^{(i,j)}\in D_{H'}$ to predict the distribution parameters $\bm{\theta} [j]$ for the $j$-th node as well as the current prediction state $\mathbf{f}_{\text{state}}^{(l)}\in\mathbb{R}^{N \cdot D_{H'}}$. In detail, on the last $L$-th layer, the network outputs instant-wise hidden state $\mathbf{X}^{(M,j)}_L\in \mathbb{R}^{D_{H'}}$ and predicts instant-wise polar distribution parameters $\bm{\theta}[j]$ applying a two-layer multi-layer perceptron (MLP) per parameter. When the distribution parameters are non-negatives (e.g., $w_j$, $\sigma^2_{d_j}$, and $\kappa_{\phi_j}$), we use softplus activation functions.
In addition, for the concentration $\kappa_{\phi_j}$, we enable the MLP to produce $\frac{1}{\kappa_\phi}$ instead of $\kappa_\phi$ since the inverse of concentration is analogous to the variance. For the parameter $\mu_{\phi_j}$, to avoid the discontinuity on angles (e.g., $\ang{0}$ and $\ang{360}$), we map $\mathbf{X}_{L}^{M,j}$ to $[x_{\phi_j}, y_{\phi_j}] \in \mathbb{R}^2$ via a two-layer MLP and then apply an atan2 activation function $\mu_{\phi_j} = \text{atan2}(y_{\phi_j}, x_{\phi_j}) \in \mathbb{R}$. 
Through iterations, the estimation network returns spatial distributions conditioned on all the possible subsequences of the given relations $r^{(1)}, ... , r^{(M)}$ in sequence. We then generate a final spatial distribution using Eq.~(\ref{eq_iter_problem}) and Eq.~(\ref{eq:weighted_sum}): $p(\bm{\theta}_{M}|r^{(1)}, ... , r^{(M)}, \Upsilon^*_{\text{sg}}) = \prod_{i=1}^{M} p_i(\bm{\theta}_i|\bm{\theta}_{i-1}, r^{(i)}, \Upsilon^*_{\text{sg}})$.

\subsubsection{Objective function}
To train the estimation network, we introduce a composite loss function $\mathcal{L}$ that is a linear combination of two loss functions, $\mathcal{L}_1$ and $\mathcal{L}_2$. $\mathcal{L}_1$ is the negative log-likelihood of the spatial distribution given the ground-truth locations $\mathbf{x}^{\text{des}}$ and ground-truth weight $w^\text{des}_j$:
\begin{align}
    \mathcal{L}_1 = - log \left(\sum^{N}_{j=1} w_j^\text{des} \cdot P(\mathbf{x}^{\text{des}}; \mu_{d_j}, \sigma_{d_j}^2, \mu_{\phi_j}, \kappa_{\phi_j})\right).
\end{align}
$\mathcal{L}_2$ is the cross-entropy loss between the predicted weight $w_j$ and the ground-truth weight $w^{\text{des}}_j$:
\begin{align}
    \mathcal{L}_2 = - \frac{1}{N} \sum^{N}_{j=1} w_j^{\text{des}} \cdot log(w_j).
\end{align}
Then, the combined loss is $\mathcal{L}=\lambda\mathcal{L}_1+(1-\lambda)\mathcal{L}_2$, where $\lambda$ is a hyperparameter. Given a composite instruction, we compute the combined loss and incrementally update the network given each relation $r^{(i)}$ in sequence.

\section{Experimental Setup}
Our experiments aim to answer the following questions: Does the proposed method improve the performance of space grounding given 1) an instruction with a single referring expression and 2) a composite instruction with multiple referring expressions? Further, can the proposed method apply to real-world tasks?

\subsection{Grounding with a Referring Expression}
We evaluate the grounding capability of inferring a location for successfully placing objects within a tabletop domain, guided by instructions containing a referring expression. We use three baseline methods with their benchmarks within the PyBullet simulator~\cite{coumans2019}. Each benchmark provides a top-down view of RGB-D images with synthesized structured instructions. Below are the training and test procedures per benchmark.

\begin{itemize}
\item \textbf{\textsc{CLIPort}}'s benchmark~\cite{shridhar2022cliport}: We use three tasks designed to pack an object ``inside" a referenced object. Task scenes contain between four to ten objects from the Google Scanned Objects~\cite{googlescanned} or primitive shapes, denoted as \textit{google} and \textit{shape}, respectively. Task instructions include objects seen during training or not, denoted as \textit{seen} and \textit{unseen}, respectively. We use instructions with referential expressions such as ``\textit{pack the bull figure in the brown box}." The assessment metric is a success score ($\in[0,1]$) reflecting the extent of relationship satisfaction between the located object volume and the desired container.
\item \textbf{\textsc{ParaGon}}'s benchmark~\cite{zhao2023differentiable}: 
This benchmark generates a dataset for the task of placing an object in the presence of semantically identical objects following one of nine directional relations: ``center,'' ``left,'' ``right,'' ``above,'' ``below,'' ``left above,'' ``left below,'' ``right above,'' and ``right below.'' The evaluation metric is a binary success score ($\in\{0,1\}$), denoting whether all predicates have been satisfied after placements. 
\item \textbf{SREM}'s benchmark~\cite{gkanatsios2023energy}: We use eight tasks designed to rearrange a colored object inside a referenced object following spatial instructions featuring one of four directional relations: ``left,'' ``right,'' ``behind,'' and ``front.'' Each scene contains between four to seven objects, as in \textsc{CLIPort} benchmark. The evaluation metric uses a success score as \textsc{CLIPort} with the most conservative threshold.
\item \textbf{LINGO-Space}'s benchmark: We introduce \textit{close-seen-colors}, \textit{close-unseen-colors}, \textit{far-seen-colors}, \textit{far-unseen-colors} tasks with new predicates, ``close" and ``far." Other setups are similar to the SREM benchmark.
\end{itemize}

For \textsc{ParaGon} benchmark, for each task, training and testing employ $400$ and $200$ scenes, respectively. Otherwise, we train models on $100$ samples, with subsequent testing performed on $200$ randomized samples.

\subsection{Grounding with Multiple Referring Expressions}
We also assess the performance of incremental grounding in the presence of multiple relations in sequence. However, benchmarks such as \textsc{CLIPort} and \textsc{ParaGon} focus on instructions with simple relations or structures. Instead, we introduce a new task labeled as \textit{composite} to illustrate better the challenge of grounding subsequent relations in environments with multiple semantically identical objects. This task assumes tabletop grounding scenarios akin to \textsc{CLIPort}. Each sample consists of $640 \times 320$ RGB-D images and synthesized referring instructions with 10-direction relations: ``left,'' ``right,'' ``above,'' ``below,'' ``left above,'' ``right above,'' ``left below,'' ``right below,'' ``close,'' and ``far''. For example, we use ``\textit{put the green ring to the left of the gray cube, the above of the gray cube, and the right of the red bowl.}" In the dataset, we randomly place two-to-seven objects, considering one-to-three independent relation phrases for training and one-to-six phrases for testing. Our benchmarks use SREM's scoring criteria below.

In this task, only one region strictly satisfies all the relations. To verify it, we use manually programmed bounding-based relation checkers. Across all tasks, we train models on $200$ samples without having semantically identical objects and repeated directional predicates. We then test on $300$ samples.

In addition, we perform evaluations with SREM's tasks designed for multiple referential instructions (i.e., ``comp-one-step-(un)seen-colors"). After grounding a target location, we compute a score reflecting the extent of relationship satisfaction; $score=\#satisfaction/|relations|$.

\subsection{Baseline Methods}
\begin{itemize}
    \item \textbf{\textsc{CLIPort}} is a language-conditioned imitation learning model that predicts pixel locations using CLIP~\cite{radford2021learning} and Transporter Networks~\cite{zeng2021transporter}. Note that we disabled its rotational augmentation except for the \textsc{CLIPort} benchmark.
    \item \textbf{\textsc{ParaGon}} is a parsing-based visual grounding model for object placement. This model generates particles as location candidates (i.e., pixels). We use the pixel point with maximum probability as a placement position. For visual feature extraction, we use the provided Mask R-CNN~\cite{he2017mask} for \textsc{ParaGon} benchmark and Grounding Dino~\cite{liu2023grounding} for the other benchmarks. 
    \item \textbf{SREM} is an instruction-guided rearrangement framework. Parsing an instruction into multiple spatial predicates, SREM's open-vocabulary visual detector \cite{jain2022bottom} grounds them to objects in input images. For LINGO-Space benchmark, we train energy-based models for new predicates. We then train the grounder with each task dataset while training the parser with each benchmark dataset following the paper. We exclude the Transporter Networks~\cite{zeng2021transporter} since we provide the ground-truth location given a bounding box. We also disable the closed-loop execution for fairness in comparison.
\end{itemize}
\section{Evaluation}

\begin{table*}[t]
\centering
\small
\setlength{\tabcolsep}{4.5pt}
\begin{tabular}{c ccc cccccccc c}
\toprule
Benchmark & \multicolumn{3}{c}{\textsc{CLIPort}} &\multicolumn{8}{c}{SREM} & \textsc{ParaGon} \\ 
\cmidrule(lr){2-4} \cmidrule(lr){5-12} \cmidrule(lr){13-13}
Task & \makecell{\small{packing}\\ \small{-seen}\\ \small{-google}} & \makecell{\small{packing}\\ \small{-unseen}\\ \small{-google}} & \makecell{\small{packing}\\ \small{-unseen}\\ \small{-shapes}} & \makecell{\small{left}\\ \small{-seen}\\ \small{-colors}} & \makecell{\small{left}\\ \small{-unseen}\\ \small{-colors}} & \makecell{\small{right}\\ \small{-seen}\\ \small{-colors}} & \makecell{\small{right}\\ \small{-unseen}\\ \small{-colors}} & \makecell{\small{behind}\\ \small{-seen}\\ \small{-colors}} & \makecell{\small{behind}\\ \small{-unseen}\\ \small{-colors}} & \makecell{\small{front}\\ \small{-seen}\\ \small{-colors}} & \makecell{\small{front}\\ \small{-unseen}\\ \small{-colors}} & \small{simple} \\ \hline
\textsc{CLIPort} & 98.0 & 97.7 & 95.5 & 88.0 & 86.5 & 90.5 & 98.5 & \textbf{99.5} & 96.0 & 98.5 & 99.0 & 38.5\\
\textsc{ParaGon} & 98.1 & \textbf{98.0} & 99.5 & 75.5 & 61.5 & 87.5 & 86.0 & 99.0 & 97.5 & 99.0 & 96.5 & \makecell{41.5 \\ \small{(85.1)}} \\
SREM & 98.7 & 97.3 & \textbf{100} & 93.0 & 94.5 & 82.0 & 81.5 & 99.0 & 98.0 & 99.0 & \textbf{100} & 42.5 \\
LINGO-Space & \textbf{99.2} & \textbf{98.0} & \textbf{100} & \textbf{99.5} & \textbf{97.5} & \textbf{99.5} & \textbf{100} & 98.5 & \textbf{98.5} & \textbf{100} & \textbf{100} & \textbf{80.0} \\ 
\bottomrule
\end{tabular}
\caption{Evaluation (success score) on $12$ benchmark tasks with a single referring expression. The $12$ tasks are from \textsc{CLIPort}, \textsc{ParaGon}, and SREM, where methods are trained and tested within each task's dataset. The score indicates how successfully each method identified a location satisfying relations in $[0,100]$. The number in the parentheses is the result of the literature.
}
\label{table:other_benchmark}
\end{table*}

\begin{table}[t]
\centering
\small
\setlength{\tabcolsep}{3pt}
\begin{tabular}{c cccc}
\toprule
 Task & \makecell{\small{close-seen}\\\small{-colors}} & \makecell{\small{close-unseen}\\\small{-colors}} & \makecell{\small{far-seen}\\\small{-colors}} & \makecell{\small{far-unseen}\\\small{-colors}} \\ \hline
\textsc{CLIPort} & 38.5 & 18.5 & 59.5 & 60.0 \\
\textsc{ParaGon} & 38.5 & 41.5 & 31.5 & 42.0 \\
SREM & \textbf{91.0} & \textbf{90.5} & 45.0 & 44.5 \\
\small{LINGO-Space} & 86.0 & 81.0 & \textbf{95.5} & \textbf{95.0} \\ 
\bottomrule
\end{tabular}
\caption{Evaluation (success score) on our $4$ benchmark tasks with new predicates: \textit{close} and \textit{far}.}
\label{table:our_benchmark_single}
\end{table}

\begin{table}[t]
\centering
\small
\setlength{\tabcolsep}{4.5pt}
\begin{tabular}{c cc c c}
\toprule
Benchmark & \textsc{ParaGon} & \multicolumn{2}{c}{SREM} & \makecell{\small{LINGO}\\\small{-Space}} \\ 
\cmidrule(lr){2-2} \cmidrule(lr){3-4} \cmidrule(lr){5-5}
Task & \makecell{\small{composi}\\\small{tional}} & \makecell{\small{comp-one}\\ \small{-step}\\ \small{-seen} \\ \small{-colors}} & \makecell{\small{comp-one}\\ \small{-step}\\ \small{-unseen} \\ \small{-colors}} & \makecell{\small{compo} \\ \small{site}}\\ \hline
\textsc{CLIPort} & 26.0 & 87.5 & 84.8 & 54.4 \\
\textsc{ParaGon} & \makecell{28.5 \\ \small{(67.9)}} & 76.4 & 77.3 & 56.0 \\
SREM & 1.5 &  93.4 & 92.1 & 42.2\\
\small{LINGO-Space} & \textbf{90.5} & \textbf{97.5} & \textbf{96.5} & \textbf{79.1}\\ 
\bottomrule
\end{tabular}
\caption{Evaluation (success score) on $4$ benchmark tasks with multiple referring expressions. The number in the parentheses is the result of the literature. 
                           }
\label{table:benchmark_composite}
\end{table}

\textbf{Grounding with a referring expression.} We analyze the grounding performance of our proposed method and baselines via $12$ benchmarks. Table~\ref{table:other_benchmark} shows our method outperforms three baseline methods, demonstrating superior performance in $11$ out of $12$ tasks. Our method consistently exhibits the highest success scores even when faced with \textit{unseen} objects, owing to its ability to incorporate embedded visual and linguistic features. However, our method exhibits a $1.0$ lower performance in the \textit{behind-seen-colors} task since our method does not account for volumes causing placement failures.
In addition, SREM fails to attain scores in the \textit{simple} task due to its dependence on a pre-defined instruction structure, in contrast to our LLM-based parser.

We extend the evaluation using novel predicates (see Table~\ref{table:our_benchmark_single}). Our method achieves consistently high scores due to its ability to represent diverse distributions corresponding to not only ``close" but also ``far," while the performance of the other methods degraded significantly. Although SREM is able to accommodate ``close" with energy-based representation, the ``far" predicates cannot be seamlessly accommodated by conventional rules or location representations.

\textbf{Grounding with multiple referring expressions.} Our method significantly improves the performance of grounding given a composite instruction. Table~\ref{table:benchmark_composite} shows our method results in superior success scores in all four tasks. Our method is a maximum of $62$ higher than the second-best approach in each benchmark since our LLM-based parser extracts relation tuples from diverse structures of instructions resolving \textit{compositional ambiguity}. \textsc{ParaGon} shows a performance degradation in their benchmark since we trained \textsc{ParaGon} with the \textit{compositional} task data only, unlike the paper setup. Although \textsc{ParaGon}'s performance record in the paper is $67.9$, the record is still $22.6$ lower than our method result. SREM shows good performance in its benchmark, training its grounder and parser with the task and benchmark datasets, respectively, following the literature. However, SREM often fails to parse the composite instructions in other datasets since it requires a specific format of composite instructions containing \textit{clause} instead of \textit{phrase} as a referring expression. 

\begin{figure}[t]
  \centering
  \includegraphics[width=\columnwidth]{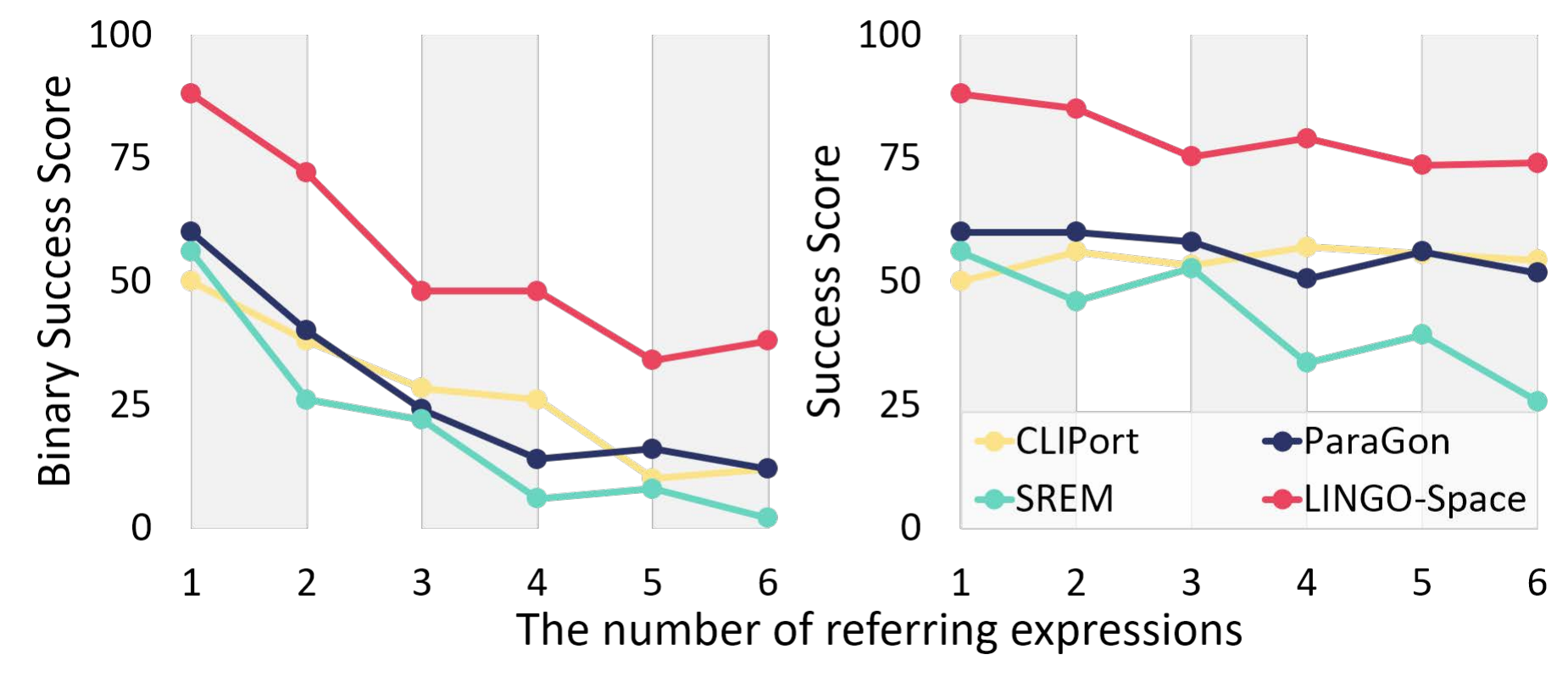}
  \caption{Grounding performance on the increasing number of expressions.  Each graph uses a distinct score metric: (left) the binary success score as \textsc{ParaGon} benchmark and (right) the success score.
  }
  \label{fig:composite_scalability}
\end{figure}

Further, as shown in Fig.~\ref{fig:composite_scalability}, our method shows the capability of handling multiple expressions with the consistently highest scores given an increasing number of referring expressions, while other approaches are going to fail to ground four to six referring expressions. This is because 1) the ``\textit{composite}" task includes multiple similar objects causing \textit{referential ambiguity} and 2) the instruction includes more predicates without using clauses. However, our method resolves them by leveraging the semantic and geometric relationships encoded in the scene graph.

\textbf{Real-world demonstrations.} We finally demonstrated the \textit{space-grounding} capability of our LINGO-Space, integrating it on a navigation framework for a quadruped robot, Spot, from Boston Dynamics. Fig.~\ref{fig_main} illustrates the language-guided navigation experiment. A human operator delivers a command, such as ``move to the front of the red box and close to the tree." Our robot generates a scene graph using a LiDAR-based bounding box detector and parses the instruction using a ChatGPT~\cite{openai2023chatgpt}. The robot then successfully identified the goal distribution, and it reached the best location we wanted.\footnote{See details on our website: https://lingo-space.github.io}
\section{Conclusion}
We introduced LINGO-Space, a language-conditioned incremental space-grounding method for composite instructions with multiple referring expressions. LINGO-space is a probabilistic grounding network leveraging a mixture of learnable polar distributions to predict probable location distributions. Our evaluation shows the effectiveness of the proposed probabilistic approach in representing desired space. Further, we demonstrate that, when coupled with the proposed LLM-guided semantic parser, our network enhances reasoning complex composite instructions with diverse referring expressions. Compared with state-of-the-art baselines, our model outperforms in grounding success score, generalizability, and scalability. We finally validate its practical usability by deploying LINGO-Space to a real-world navigation task running a quadruped robot. 

\section{Acknowledgements}
\bigskip
\noindent This work was partly supported by Institute of Information \& communications Technology Planning \& Evaluation (IITP) grant funded by the Korea government (MSIT) (No.2022-0-00311), National Research Foundation of Korea (NRF) grants funded by the Korea government (MSIT) (No.2021R1A4A3032834 and 2021R1C1C1004368).

\bibliography{aaai24}

\begin{thebibliography}{44}
\providecommand{\natexlab}[1]{#1}

\bibitem[{Brohan et~al.(2023)Brohan, Chebotar, Finn, Hausman, Herzog, Ho, Ibarz, Irpan, Jang, Julian et~al.}]{brohan2023can}
Brohan, A.; Chebotar, Y.; Finn, C.; Hausman, K.; Herzog, A.; Ho, D.; Ibarz, J.; Irpan, A.; Jang, E.; Julian, R.; et~al. 2023.
\newblock Do as i can, not as i say: Grounding language in robotic affordances.
\newblock In \emph{Proceedings of the Conference on Robot Learning (CoRL)}, 287--318. PMLR.

\bibitem[{Coumans and Bai(2016)}]{coumans2019}
Coumans, E.; and Bai, Y. 2016.
\newblock PyBullet, a Python module for physics simulation for games, robotics and machine learning.
\newblock \url{http://pybullet.org}.

\bibitem[{Dong and Lapata(2016)}]{dong2016language}
Dong, L.; and Lapata, M. 2016.
\newblock Language to logical form with neural attention.
\newblock In \emph{Proceedings of the Association for Computational Linguistics (ACL)}, 33--43.

\bibitem[{Downs et~al.(2022)Downs, Francis, Koenig, Kinman, Hickman, Reymann, McHugh, and Vanhoucke}]{googlescanned}
Downs, L.; Francis, A.; Koenig, N.; Kinman, B.; Hickman, R.; Reymann, K.; McHugh, T.~B.; and Vanhoucke, V. 2022.
\newblock Google scanned objects: A high-quality dataset of 3D scanned household items.
\newblock In \emph{Proceedings of the IEEE International Conference on Robotics and Automation (ICRA)}, 2553--2560. IEEE.

\bibitem[{Driess et~al.(2023)Driess, Xia, Sajjadi, Lynch, Chowdhery, Ichter, Wahid, Tompson, Vuong, Yu, Huang, Chebotar, Sermanet, Duckworth, Levine, Vanhoucke, Hausman, Toussaint, Greff, Zeng, Mordatch, and Florence}]{driess2023palme}
Driess, D.; Xia, F.; Sajjadi, M. S.~M.; Lynch, C.; Chowdhery, A.; Ichter, B.; Wahid, A.; Tompson, J.; Vuong, Q.; Yu, T.; Huang, W.; Chebotar, Y.; Sermanet, P.; Duckworth, D.; Levine, S.; Vanhoucke, V.; Hausman, K.; Toussaint, M.; Greff, K.; Zeng, A.; Mordatch, I.; and Florence, P. 2023.
\newblock PaLM-E: An embodied multimodal language model.
\newblock In \emph{Proceedings of the International Conference on Machine Learning (ICML)}. PMLR.

\bibitem[{Gkanatsios et~al.(2023)Gkanatsios, Jain, Xian, Zhang, Atkeson, and Fragkiadaki}]{gkanatsios2023energy}
Gkanatsios, N.; Jain, A.; Xian, Z.; Zhang, Y.; Atkeson, C.~G.; and Fragkiadaki, K. 2023.
\newblock Energy-based models are zero-shot planners for compositional scene rearrangement.
\newblock In \emph{Proceedings of Robotics: Science and Systems (RSS)}.

\bibitem[{Guadarrama et~al.(2013)Guadarrama, Riano, Golland, Go, Jia, Klein, Abbeel, Darrell et~al.}]{guadarrama2013grounding}
Guadarrama, S.; Riano, L.; Golland, D.; Go, D.; Jia, Y.; Klein, D.; Abbeel, P.; Darrell, T.; et~al. 2013.
\newblock Grounding spatial relations for human-robot interaction.
\newblock In \emph{Proceedings of the IEEE/RSJ International Conference on Intelligent Robots and Systems (IROS)}, 1640--1647. IEEE.

\bibitem[{Hatori et~al.(2018)Hatori, Kikuchi, Kobayashi, Takahashi, Tsuboi, Unno, Ko, and Tan}]{hatori2018interactively}
Hatori, J.; Kikuchi, Y.; Kobayashi, S.; Takahashi, K.; Tsuboi, Y.; Unno, Y.; Ko, W.; and Tan, J. 2018.
\newblock Interactively picking real-world objects with unconstrained spoken language instructions.
\newblock In \emph{Proceedings of the IEEE International Conference on Robotics and Automation (ICRA)}, 3774--3781. IEEE.

\bibitem[{He et~al.(2017)He, Gkioxari, Doll{\'a}r, and Girshick}]{he2017mask}
He, K.; Gkioxari, G.; Doll{\'a}r, P.; and Girshick, R. 2017.
\newblock Mask r-cnn.
\newblock In \emph{Proceedings of the International Conference on Computer Vision (ICCV)}, 2961--2969.

\bibitem[{Howard et~al.(2022)Howard, Stump, Fink, Arkin, Paul, Park, Roy et~al.}]{howard2022intelligence}
Howard, T.; Stump, E.; Fink, J.; Arkin, J.; Paul, R.; Park, D.; Roy, S.; et~al. 2022.
\newblock An intelligence architecture for grounded language communication with field robots.
\newblock \emph{Field Robotics}, 468--512.

\bibitem[{Howard, Tellex, and Roy(2014)}]{howard2014natural}
Howard, T.~M.; Tellex, S.; and Roy, N. 2014.
\newblock A natural language planner interface for mobile manipulators.
\newblock In \emph{Proceedings of the IEEE International Conference on Robotics and Automation (ICRA)}, 6652--6659. IEEE.

\bibitem[{Hu et~al.(2020)Hu, Liu, Gomes, Zitnik, Liang, Pande, and Leskovec}]{hu2020strategies}
Hu, W.; Liu, B.; Gomes, J.; Zitnik, M.; Liang, P.; Pande, V.; and Leskovec, J. 2020.
\newblock Strategies for pre-training graph neural networks.
\newblock In \emph{Proceedings of the International Conference on Learning Representation (ICLR)}.

\bibitem[{Huang et~al.(2023)Huang, Xia, Xiao, Chan, Liang, Florence, Zeng, Tompson, Mordatch, Chebotar et~al.}]{huang2022inner}
Huang, W.; Xia, F.; Xiao, T.; Chan, H.; Liang, J.; Florence, P.; Zeng, A.; Tompson, J.; Mordatch, I.; Chebotar, Y.; et~al. 2023.
\newblock Inner monologue: Embodied reasoning through planning with language models.
\newblock In \emph{Proceedings of the Conference on Robot Learning (CoRL)}, 1769--1782. PMLR.

\bibitem[{Jain et~al.(2022)Jain, Gkanatsios, Mediratta, and Fragkiadaki}]{jain2022bottom}
Jain, A.; Gkanatsios, N.; Mediratta, I.; and Fragkiadaki, K. 2022.
\newblock Bottom up top down detection transformers for language grounding in images and point clouds.
\newblock In \emph{Proceedings of the European Conference on Computer Vision (ECCV)}, 417--433. Springer.

\bibitem[{Jain et~al.(2023)Jain, Chhangani, Tiwari, Krishna, and Gandhi}]{jain2023ground}
Jain, K.; Chhangani, V.; Tiwari, A.; Krishna, K.~M.; and Gandhi, V. 2023.
\newblock Ground then navigate: Language-guided navigation in dynamic scenes.
\newblock In \emph{Proceedings of the IEEE International Conference on Robotics and Automation (ICRA)}, 4113--4120. IEEE.

\bibitem[{Kartmann et~al.(2020)Kartmann, Zhou, Liu, Paus, and Asfour}]{kartmann2020representing}
Kartmann, R.; Zhou, Y.; Liu, D.; Paus, F.; and Asfour, T. 2020.
\newblock Representing spatial object relations as parametric polar distribution for scene manipulation based on verbal commands.
\newblock In \emph{Proceedings of the IEEE/RSJ International Conference on Intelligent Robots and Systems (IROS)}, 8373--8380. IEEE.

\bibitem[{Kim et~al.(2023)Kim, Kim, Jang, Song, Choi, and Park}]{kim2023sggnet}
Kim, D.; Kim, Y.; Jang, J.; Song, M.; Choi, W.; and Park, D. 2023.
\newblock SGGNet$^2$: Speech-scene graph grounding network for speech-guided navigation.
\newblock In \emph{Proceedings of the IEEE International Conference on Robot and Human Interactive Communication (RO-MAN)}, 1648--1654. IEEE.

\bibitem[{Liang et~al.(2023)Liang, Huang, Xia, Xu, Hausman, Ichter, Florence, and Zeng}]{liang2023code}
Liang, J.; Huang, W.; Xia, F.; Xu, P.; Hausman, K.; Ichter, B.; Florence, P.; and Zeng, A. 2023.
\newblock Code as policies: Language model programs for embodied control.
\newblock In \emph{Proceedings of the IEEE International Conference on Robotics and Automation (ICRA)}, 9493--9500. IEEE.

\bibitem[{Liu et~al.(2022)Liu, Yang, Idrees, Liang, Schornstein, Tellex, and Shah}]{liu2023lang2ltl}
Liu, J.~X.; Yang, Z.; Idrees, I.; Liang, S.; Schornstein, B.; Tellex, S.; and Shah, A. 2022.
\newblock Lang2LTL: Translating natural language commands to temporal robot task specification.
\newblock In \emph{The Workshop on Language and Robotics at Conference on robot learning}.

\bibitem[{Liu et~al.(2023)Liu, Zeng, Ren, Li, Zhang, Yang, Li, Yang, Su, Zhu et~al.}]{liu2023grounding}
Liu, S.; Zeng, Z.; Ren, T.; Li, F.; Zhang, H.; Yang, J.; Li, C.; Yang, J.; Su, H.; Zhu, J.; et~al. 2023.
\newblock Grounding dino: Marrying dino with grounded pre-training for open-set object detection.
\newblock \emph{arXiv preprint arXiv:2303.05499}.

\bibitem[{Matuszek et~al.(2012)Matuszek, FitzGerald, Zettlemoyer, Bo, and Fox}]{matuszek2012joint}
Matuszek, C.; FitzGerald, N.; Zettlemoyer, L.; Bo, L.; and Fox, D. 2012.
\newblock A joint model of language and perception for grounded attribute learning.
\newblock In \emph{Proceedings of the International Conference on Machine Learning (ICML)}, 1435--1442.

\bibitem[{Mees, Borja-Diaz, and Burgard(2023)}]{mees23hulc2}
Mees, O.; Borja-Diaz, J.; and Burgard, W. 2023.
\newblock Grounding language with visual affordances over unstructured data.
\newblock In \emph{Proceedings of the IEEE International Conference on Robotics and Automation (ICRA)}, 11576--11582. IEEE.

\bibitem[{Mees et~al.(2020)Mees, Emek, Vertens, and Burgard}]{mees2020learning}
Mees, O.; Emek, A.; Vertens, J.; and Burgard, W. 2020.
\newblock Learning object placements for relational instructions by hallucinating scene representations.
\newblock In \emph{Proceedings of the IEEE International Conference on Robotics and Automation (ICRA)}, 94--100. IEEE.

\bibitem[{Mildenhall et~al.(2021)Mildenhall, Srinivasan, Tancik, Barron, Ramamoorthi, and Ng}]{mildenhall2021nerf}
Mildenhall, B.; Srinivasan, P.~P.; Tancik, M.; Barron, J.~T.; Ramamoorthi, R.; and Ng, R. 2021.
\newblock Nerf: Representing scenes as neural radiance fields for view synthesis.
\newblock \emph{Communications of the ACM}, 65(1): 99--106.

\bibitem[{Namasivayam et~al.(2023)Namasivayam, Singh, Bindal, Tuli, Agrawal, Jain, Singla, and Paul}]{namasivayam2023learning}
Namasivayam, K.; Singh, H.; Bindal, V.; Tuli, A.; Agrawal, V.; Jain, R.; Singla, P.; and Paul, R. 2023.
\newblock Learning neuro-symbolic programs for language guided robot manipulation.
\newblock In \emph{Proceedings of the IEEE International Conference on Robotics and Automation (ICRA)}, 7973--7980. IEEE.

\bibitem[{OpenAI(2023)}]{openai2023chatgpt}
OpenAI. 2023.
\newblock ChatGPT (Aug 14 version).
\newblock \url{https://chat.openai.com/chat}.
\newblock Large language model.

\bibitem[{Paul et~al.(2018)Paul, Arkin, Aksaray, Roy, and Howard}]{paul2018efficient}
Paul, R.; Arkin, J.; Aksaray, D.; Roy, N.; and Howard, T.~M. 2018.
\newblock Efficient grounding of abstract spatial concepts for natural language interaction with robot platforms.
\newblock \emph{International Journal of Robotics Research}, 37(10): 1269--1299.

\bibitem[{Paxton et~al.(2022)Paxton, Xie, Hermans, and Fox}]{paxton2022predicting}
Paxton, C.; Xie, C.; Hermans, T.; and Fox, D. 2022.
\newblock Predicting stable configurations for semantic placement of novel objects.
\newblock In \emph{Proceedings of the Conference on Robot Learning (CoRL)}, 806--815. PMLR.

\bibitem[{Radford et~al.(2021)Radford, Kim, Hallacy, Ramesh, Goh, Agarwal, Sastry, Askell, Mishkin, Clark et~al.}]{radford2021learning}
Radford, A.; Kim, J.~W.; Hallacy, C.; Ramesh, A.; Goh, G.; Agarwal, S.; Sastry, G.; Askell, A.; Mishkin, P.; Clark, J.; et~al. 2021.
\newblock Learning transferable visual models from natural language supervision.
\newblock In \emph{Proceedings of the International Conference on Machine Learning (ICML)}, 8748--8763. PMLR.

\bibitem[{Ramp{\'a}{\v{s}}ek et~al.(2022)Ramp{\'a}{\v{s}}ek, Galkin, Dwivedi, Luu, Wolf, and Beaini}]{rampavsek2022recipe}
Ramp{\'a}{\v{s}}ek, L.; Galkin, M.; Dwivedi, V.~P.; Luu, A.~T.; Wolf, G.; and Beaini, D. 2022.
\newblock Recipe for a general, powerful, scalable graph transformer.
\newblock \emph{Conference on Neural Information Processing Systems (NeurIPS)}, 35: 14501--14515.

\bibitem[{Ren et~al.(2023)Ren, Govil, Yang, Narasimhan, and Majumdar}]{ren2023leveraging}
Ren, A.~Z.; Govil, B.; Yang, T.-Y.; Narasimhan, K.~R.; and Majumdar, A. 2023.
\newblock Leveraging language for accelerated learning of tool manipulation.
\newblock In \emph{Proceedings of the Conference on Robot Learning (CoRL)}, 1531--1541. PMLR.

\bibitem[{Roy et~al.(2019)Roy, Noseworthy, Paul, Park, and Roy}]{roy2019leveraging}
Roy, S.; Noseworthy, M.; Paul, R.; Park, D.; and Roy, N. 2019.
\newblock Leveraging past references for robust language grounding.
\newblock In \emph{Proceedings of the Association for Computational Linguistics (ACL)}, 430--440.

\bibitem[{Shah et~al.(2022)Shah, Osi{\'n}ski, Levine et~al.}]{shah2022lmnav}
Shah, D.; Osi{\'n}ski, B.; Levine, S.; et~al. 2022.
\newblock {LM}-Nav: Robotic navigation with large pre-trained models of language, vision, and action.
\newblock In \emph{Proceedings of the Conference on Robot Learning (CoRL)}, 492--504. PMLR.

\bibitem[{Shridhar, Manuelli, and Fox(2022)}]{shridhar2022cliport}
Shridhar, M.; Manuelli, L.; and Fox, D. 2022.
\newblock Cliport: What and where pathways for robotic manipulation.
\newblock In \emph{Proceedings of the Conference on Robot Learning (CoRL)}, 894--906. PMLR.

\bibitem[{Shridhar, Mittal, and Hsu(2020)}]{shridhar2020ingress}
Shridhar, M.; Mittal, D.; and Hsu, D. 2020.
\newblock INGRESS: Interactive visual grounding of referring expressions.
\newblock \emph{International Journal of Robotics Research}, 39(2-3): 217--232.

\bibitem[{Singh et~al.(2023)Singh, Blukis, Mousavian, Goyal, Xu, Tremblay, Fox, Thomason, and Garg}]{singh2023progprompt}
Singh, I.; Blukis, V.; Mousavian, A.; Goyal, A.; Xu, D.; Tremblay, J.; Fox, D.; Thomason, J.; and Garg, A. 2023.
\newblock Progprompt: Generating situated robot task plans using large language models.
\newblock In \emph{Proceedings of the IEEE International Conference on Robotics and Automation (ICRA)}, 11523--11530. IEEE.

\bibitem[{Song et~al.(2023)Song, Wu, Washington, Sadler, Chao, and Su}]{song2023llm}
Song, C.~H.; Wu, J.; Washington, C.; Sadler, B.~M.; Chao, W.-L.; and Su, Y. 2023.
\newblock Llm-planner: Few-shot grounded planning for embodied agents with large language models.
\newblock In \emph{Proceedings of the International Conference on Computer Vision (ICCV)}, 2998--3009.

\bibitem[{Stopp et~al.(1994)Stopp, Gapp, Herzog, Laengle, and Lueth}]{stopp1994utilizing}
Stopp, E.; Gapp, K.-P.; Herzog, G.; Laengle, T.; and Lueth, T.~C. 1994.
\newblock Utilizing spatial relations for natural language access to an autonomous mobile robot.
\newblock In \emph{Proceedings of the German Annual Conference on Artificial Intelligence}, 39--50. Springer.

\bibitem[{Tan, Ju, and Liu(2014)}]{tan2014grounding}
Tan, J.; Ju, Z.; and Liu, H. 2014.
\newblock Grounding spatial relations in natural language by fuzzy representation for human-robot interaction.
\newblock In \emph{Proceedings of the IEEE International Conference on Fuzzy Systems (FUZZ-IEEE)}, 1743--1750. IEEE.

\bibitem[{Tellex et~al.(2011)Tellex, Kollar, Dickerson, Walter, Banerjee, Teller, and Roy}]{tellex2011understanding}
Tellex, S.; Kollar, T.; Dickerson, S.; Walter, M.; Banerjee, A.; Teller, S.; and Roy, N. 2011.
\newblock Understanding natural language commands for robotic navigation and mobile manipulation.
\newblock In \emph{Proceedings of the National Conference on Artificial Intelligence (AAAI)}, volume~25, 1507--1514. AAAI Press.

\bibitem[{Vaswani et~al.(2017)Vaswani, Shazeer, Parmar, Uszkoreit, Jones, Gomez, Kaiser, and Polosukhin}]{vaswani2017attention}
Vaswani, A.; Shazeer, N.; Parmar, N.; Uszkoreit, J.; Jones, L.; Gomez, A.~N.; Kaiser, {\L}.; and Polosukhin, I. 2017.
\newblock Attention is all you need.
\newblock \emph{Conference on Neural Information Processing Systems (NeurIPS)}, 30.

\bibitem[{Venkatesh et~al.(2021)Venkatesh, Biswas, Upadrashta, Srinivasan, Talukdar, and Amrutur}]{venkatesh2021spatial}
Venkatesh, S.~G.; Biswas, A.; Upadrashta, R.; Srinivasan, V.; Talukdar, P.; and Amrutur, B. 2021.
\newblock Spatial reasoning from natural language instructions for robot manipulation.
\newblock In \emph{Proceedings of the IEEE International Conference on Robotics and Automation (ICRA)}, 11196--11202. IEEE.

\bibitem[{Zeng et~al.(2021)Zeng, Florence, Tompson, Welker, Chien, Attarian, Armstrong, Krasin, Duong, Sindhwani et~al.}]{zeng2021transporter}
Zeng, A.; Florence, P.; Tompson, J.; Welker, S.; Chien, J.; Attarian, M.; Armstrong, T.; Krasin, I.; Duong, D.; Sindhwani, V.; et~al. 2021.
\newblock Transporter networks: Rearranging the visual world for robotic manipulation.
\newblock In \emph{Proceedings of the Conference on Robot Learning (CoRL)}, 726--747. PMLR.

\bibitem[{Zhao, Lee, and Hsu(2023)}]{zhao2023differentiable}
Zhao, Z.; Lee, W.~S.; and Hsu, D. 2023.
\newblock Differentiable parsing and visual grounding of natural language instructions for object placement.
\newblock In \emph{Proceedings of the IEEE International Conference on Robotics and Automation (ICRA)}, 11546--11553. IEEE.

\end{thebibliography}

\end{document}